\documentclass[sigconf,nonacm]{acmart}
\renewcommand\footnotetextcopyrightpermission[1]{} 
\usepackage{natbib}
\usepackage{graphicx}
\graphicspath{ {./images/} }
\usepackage[font={footnotesize},skip=0.2pt]{caption}
\usepackage{setspace}
\usepackage{subcaption}
\usepackage{siunitx}
\newcolumntype{d}{S[
    input-open-uncertainty=,
    input-close-uncertainty=,
    parse-numbers = false,
    table-align-text-pre=false,
    table-align-text-post=false
 ]}
\AtBeginDocument{%
  \providecommand\BibTeX{{%
    \normalfont B\kern-0.5em{\scshape i\kern-0.25em b}\kern-0.8em\TeX}}}

\newcommand{\blfootnote}[1]{\begingroup\renewcommand{\thefootnote}{}\footnote{#1}\addtocounter{footnote}{-1}\endgroup}
\begin{document}

\title{Estimating Supply Incrementality in Two-sided Marketplaces: A Causal Machine Learning Approach}

\author{Yufei Wu}
\affiliation{%
  \institution{Airbnb, Inc.}
  \city{San Francisco}
  \state{California}
  \country{USA}
}
\email{yufei.wu@airbnb.com}

\author{Daniel Schmierer}
\affiliation{%
  \institution{Airbnb, Inc.}
  \city{San Francisco}
  \state{California}
  \country{USA}
}
\email{daniel.schmierer@airbnb.com}

\author{Dan Zylberglejd}
\affiliation{%
  \institution{Airbnb, Inc.}
  \city{San Francisco}
  \state{California}
  \country{USA}
}
\email{dan.zylberglejd@airbnb.com}


\begin{abstract}
In two-sided marketplaces with heterogeneous products, it is important to understand the causal relationship between additional supply and marketplace outcomes, such as the total quantity transacted or transaction value in the marketplace. This paper studies a causal machine learning approach to estimating this relationship across product segments. We use the Airbnb marketplace as an example, focusing on the impact of additional listing supply on total bookings, but the methodology applies to other two-sided marketplaces. Our approach combines double/debiased machine learning with a hierarchical Bayesian framework that leverages pre-existing knowledge as priors.  We construct tractable and informative features for the model by leveraging measures of product segment similarity from the geospatial literature.  We find that such a model provides plausible estimates of the marketplace returns to additional supply and strong out of sample performance. 
\end{abstract}


\maketitle
\blfootnote{\textcopyright\ 2025 Airbnb, Inc. All rights reserved.}

\section{Introduction} \label{Intro}
Two-sided marketplaces face the fundamental challenge of balancing supply and demand across heterogeneous product segments. This is critical for marketplaces to optimize product features to better serve both sides of the marketplace. For example, Airbnb provides tools that help hosts to choose and set competitive prices, in order to best meet the demand from guests (\citet{ye2018customized}). 

As a critical piece of the general puzzle for balancing supply and demand, it is important to understand the incremental value of additional supply for different listing segments. We will refer to this as supply incrementality. An Airbnb blog post by \citet{sanchez2025airbnb} discusses supply incrementality as part of estimating the lifetime value for each listing. The Venn diagram in their Figure 1 illustrates the concept of supply incrementality, or by how much certain listings actually grow total bookings instead of cannibalizing bookings from pre-existing listings. The pink circle represents total bookings with the pre-existing listings, while the blue circle represents bookings with certain new listings. The overlapping section represents cannibalization — bookings that shifted from other listings to the new supply — while the non-overlapping portion of the blue circle reflects the new bookings that are generated by the new supply. Higher supply incrementality of a listing segment means there had been insufficient supply to meet demand, and the additional supply helped meet the demand from guests. 

\begin{figure}[h]
  \caption{Figure 1 from Sanchez Martinez et al. (2025)}
  \centering
  \includegraphics[width=1\linewidth]{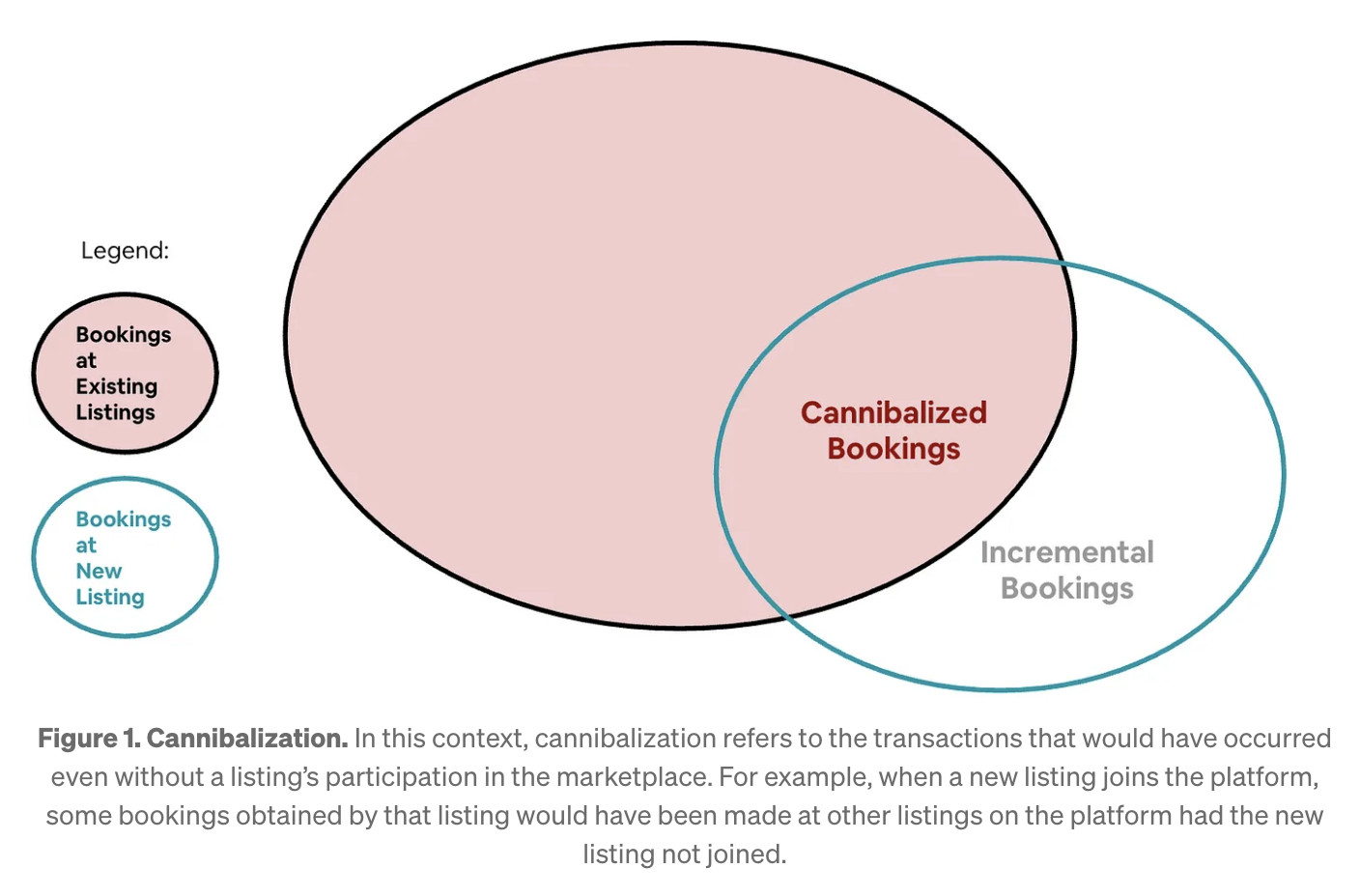}
  \label{fig:incr}
\end{figure}

\citet{sanchez2025airbnb} also noted examples of why such work helps serve the community more effectively, such as pinpointing listing segments that have an opportunity to get more bookings and identifying which marketing initiatives bring the most value to the community.  

As important as it is to measure supply incrementality, it presents significant challenges. The core challenge is that we don’t observe the counterfactual: what would have happened in the marketplace if certain supply had not been available — it’s not practical to run experiments that randomize supply availability, and observational data is confounded by endogeneity and substitution effects.

\begin{itemize}
    \item Experiments have a long history in online marketplaces and, when done correctly, are still seen as the gold standard for causal inference. However, randomly assigning supply is typically infeasible, so we need to use observational data instead and isolate the impact of supply on bookings. 

    \item The challenge with using observational data is that listing supply and guest demand are endogenous — more hosts might choose to list their places in listing segments facing high demand, creating correlation between supply and bookings that will generally confound any causal relationship between supply and marketplace outcomes. 

    \item Substitution between different listings also impose challenges in measuring the right supply incrementality instead of cannibalization. When a new listing becomes available, it may get some bookings from guests who would otherwise be booked with other listings, and we need to isolate the true incrementality from that substitution effect. As \citet{sanchez2025airbnb} also discussed, "Accounting for incrementality is challenging because we never observe the ground truth. While we observe how many bookings are made per listing, we cannot tell which bookings are incremental and which bookings are cannibalized from other listings." 
     
    \item Finally, we need to not only understand supply incrementality overall, but also how that varies across different listing segments. In general, the impact of additional supply likely varies considerably across product segments, requiring flexible models that can capture this heterogeneity and provide heterogeneous treatment effect estimates. 

\end{itemize}

To address these challenges, we combine a causal machine learning approach and a hierarchical Bayesian framework, incorporating measures of product similarity based on the geospatial literature.

\begin{itemize}
    \item To mitigate endogeneity and focus on isolating the causal impact of supply, we incorporate rich signals on demand for and supply of each listing segment into a Double Machine Learning framework. In the first stage, we use these signals to explain historical variation in supply and bookings; in the second stage, we use the residuals from the first stage models to isolate the impact of supply on bookings, holding everything else the same. 

    \item Our model requires focusing on features that drive supply and demand in a given (narrow) listing segment. As such, we need to focus just on the listing segments that are most relevant to the supply and demand – incorporating data on distant listing segments just introduces additional noise. We therefore follow the geospatial literature and incorporate measures of similarity between listing segments to improve the predictive power of the first stage and reduce variance of the heterogeneous treatment effect estimates in the second stage. 
    
    \item Finally, we augment this approach with a hierarchical Bayesian framework that leverages pre-existing knowledge as priors, while leveraging new data to update our belief where there are strong data signals that deviate from the prior.

    \item In the second stage of the Double ML model, we estimate heterogeneous treatment effects as a function of important marketplace and product characteristics. These allow us to understand how various factors influence the incrementality of additional supply in different listing segments. 

\end{itemize}
We find that such models are able to provide plausible estimates of the marketplace returns to additional supply. Even though our discussion is focused on the Airbnb marketplace, the methodology we develop has broad applications to other two-sided marketplaces where understanding the causal impact of supply changes is valuable. 

\section{Methodology} \label{Method}
\subsection{Model Functional Form}
\citet{sanchez2025airbnb} laid the groundwork for estimating supply incrementality at Airbnb, which is based on a Cobb-Douglas model. The Cobb-Douglas model is a widely used production function that relates output to the inputs of capital and labor (\citet{douglas1976cobb}) and has been used more recently as a matching function for matching workers with job vacancies (\citet{petrongolo2001looking}). In the context of marketplace supply incrementality, it models the relationship between supply, demand, and marketplace outcomes (bookings) with constant outcome elasticities for each input. The referenced blog post applies this model to estimate how increases in supply of listings affect total bookings, with incrementality parameters indicating the percentage change in bookings for a percentage change in supply.

There are also other structural models of demand and supply in marketplaces with heterogeneous products. For example, as \citet{gandhi2021empirical} summarizes, structural models typically employ discrete choice frameworks like Berry, Levinsohn, Pakes (BLP) models (\citet{berry1995automobile}) or random coefficient logit models to capture consumer preferences across differentiated products. These models often specify utility functions, impose assumptions about substitution patterns, and estimate demand parameters using methods like generalized method of moments.

Our approach imposes less strict functional form assumptions on the relationship between supply, demand, and marketplace outcomes. We take this approach for several key reasons:
\begin{enumerate}
    \item The Double ML approach leverages ML models to allow for more flexible relationships between variables (\citet{chernozhukov2018double}).

    \item It provides better control for confounding factors that affect both supply and bookings.

    \item It can accommodate complex non-linear relationships and interactions between variables without requiring pre-specification of the functional form.

    \item It handles high-dimensional data more effectively, incorporating many potential predictors.

    \item This type of approach is also well suited for heterogeneous treatment effects, capturing how the impact of additional supply varies across different listing segments (\citet{kennedy2022towards}).
\end{enumerate}

\subsection{Double Machine Learning Model Set-up}
The Double ML approach works by first predicting both supply and bookings using other features in the first stage, and then regressing residual bookings on residual supply in the second stage. Intuitively, we try to focus on the causal impact of supply on bookings, after accounting for the potential influence of other factors such as demand-supply tightness, listing attributes, etc.

Let $g$ denote a broad grouping of listing segments, $j$ denote listing segments, and $t$ denote time period. We construct the following metrics at the product segment $\times$ time period level, from which we can construct metrics to estimate the model. The broad grouping of listing segments ($g$) is defined in a way to minimize cannibalization, the same as the listing segments considered by \citet{sanchez2025airbnb}. However, for more granular listing segments ($j$), there could be overlap in demand across listing segments, making it important to consider the substitution between different listing segments. 

Let $Y$ denote the outcome, bookings, $S$ denote treatment, supply, and $X$ denote covariates, including demand patterns, listing features, trend, and seasonality. In the first stage of the Double ML model, we use all the covariates to help explain the observed variation in supply and bookings, so we can focus on the remaining unexplained variation for the second stage. In other words, we first predict $Y$ and $S$ based on the same $X$ on the right hand side, so we can later estimate the impact of residual $S$ on residual $Y$, isolating the impact of supply on bookings, holding everything else constant.

A key property of Double ML is its "double robustness" - the estimator remains consistent and asymptotically normal even if one of the first-stage models is misspecified, as long as the other model is correctly specified. As \citet{chernozhukov2018double} note, "Both components need to be explicitly estimated to construct a debiased estimator. Using both components, however, yields an extra degree of robustness" (p. C5). This robustness property is particularly valuable for measuring supply incrementality because it helps separate the true causal effect from correlation due to common demand drivers, even in cases where either the supply or bookings model might be imperfectly specified.

For model selection for the first-stage models, we evaluated various algorithms based on validation data performance. Among the considered models, \href{http://lightgbm.readthedocs.io/}{LightGBM} demonstrated better out-of-sample fit for both predicting outcome $Y$ and treatment $S$, making it our preferred choice for the first-stage predictions in the Double ML framework.

\subsection{Time Series Data Split}
For model estimation and validation, we employ a time series split approach rather than traditional random cross-validation to avoid overfitting due to shared seasonality across listing segments. This choice is crucial when working with time series data like ours, where random splitting would violate the temporal structure and lead to data leakage problems. Time series data exhibits temporal dependencies that make observations from nearby time periods correlated, and random splitting would allow models to "peek into the future" during training, resulting in overly optimistic performance estimates (\citet{bergmeir2018note}).

\subsection{Accounting for Substitution Between Listing Segments}

As mentioned above, while we can abstract away from substitution between the listing groups ($g$), we need to account for potential substitution between listing segments ($j$). Therefore, for predicting $Y$ and $S$ in the first stage for listing segment $j$, besides the covariates $X$ in listing segment $j$ ($X_{g,j,t}$), we also account for the covariates $X$ in other listing segments $j$ within the same group $g$ ($X_{g,-j,t}$), as well as available supply in those other listing segments ($S_{g,-j,t}$).

\begin{equation}
Y_{g,j,t} = \alpha(t) + \tilde{\beta} S_{g,-j,t} + \lambda X_{g,j,t} + \tilde{\lambda} X_{g,-j,t} + \epsilon_{g,j,t}
\label{eq:first_stage_y}
\end{equation}

\begin{equation}
S_{g,j,t} = \delta(t) + \tilde{\theta}S_{g,-j,t} + \omega X_{g,j,t} + \tilde{\omega} X_{g,-j,t} + u_{g,j,t}
\label{eq:first_stage_s}
\end{equation}

Rather than treating all listing segments independently, we leverage these relationships through a geospatial approach that incorporates similarity measures between listing segments (\citet{credit2024structured}). The geospatial approach leverages the insight that listing segments that are most similar to each other are likely to have similar patterns of demand and supply dynamics due to underlying spatial and feature-based dependencies. Incorporating measures of similarity between listing segments results in improved predictions in the first stage model and reduced variance in the second stage treatment effect estimation.

\subsection{Second Stage: Understand How Supply Incrementality Varies Across Listing Segments}

In the second stage, we parameterize the heterogenous treatment effect as a function of $X$. We adopt a Bayesian approach to leverage this pre-existing knowledge (\citet{sanchez2025airbnb}) as priors and only update our belief where new data signals strongly suggest deviations from the prior. 

Specifically, we estimate a hierarchical model as follows, formulating the heterogenous treatment effect as a combination of listing group level random effects and the interaction effects reflecting how supply incrementality varies with $X$ in a listing segment as well as the most similar listing segments in the same group.

\begin{equation}
Y^{\text{Residual}}_{g,j,t} = \theta_{g,j} S^{\text{Residual}}_{g, j, t} + \epsilon_{g, j, t}
\label{eq:second_stage}
\end{equation}

\begin{equation}
\theta_{g,j} = \alpha_{g} +  \lambda X_{g,j,t} + \tilde{\lambda} X_{g,-j,t}
\label{eq:heterogeneous_effect}
\end{equation}

In this specification, $\alpha_g \sim N(E_g, \sigma_\alpha^2)$ is a random effect reflecting listing group level base supply incrementality, using listing group level supply incrementality $E_g$ from the previous model to inform prior distributions. The other two terms reflect the interaction effects with the covariates in the listing segment as well as in its nearest neighbors, with the parameters following uninformative priors $\lambda \sim N(0, \sigma_\lambda^2)$ and $\tilde{\lambda} \sim N(0, \sigma_{\tilde{\lambda}}^2)$.

\section{Findings on Supply Incrementality} \label{Findings}
Our findings demonstrate that the combined approach of causal machine learning with geospatial relationships and hierarchical Bayesian modeling provides plausible and robust estimates of supply incrementality across heterogeneous listing segments in the Airbnb marketplace. The methodology successfully captures the varying impact of additional supply across different listing segments, identifying segments where additional supply would bring the most value to the community. 

\subsection{Prior vs Posterior of the Heterogeneous Treatment Effects}

\begin{figure}[h]
  \caption{Prior vs Posterior for Interaction Coefficient Estimates}
  \centering
  \includegraphics[width=1.06\linewidth]{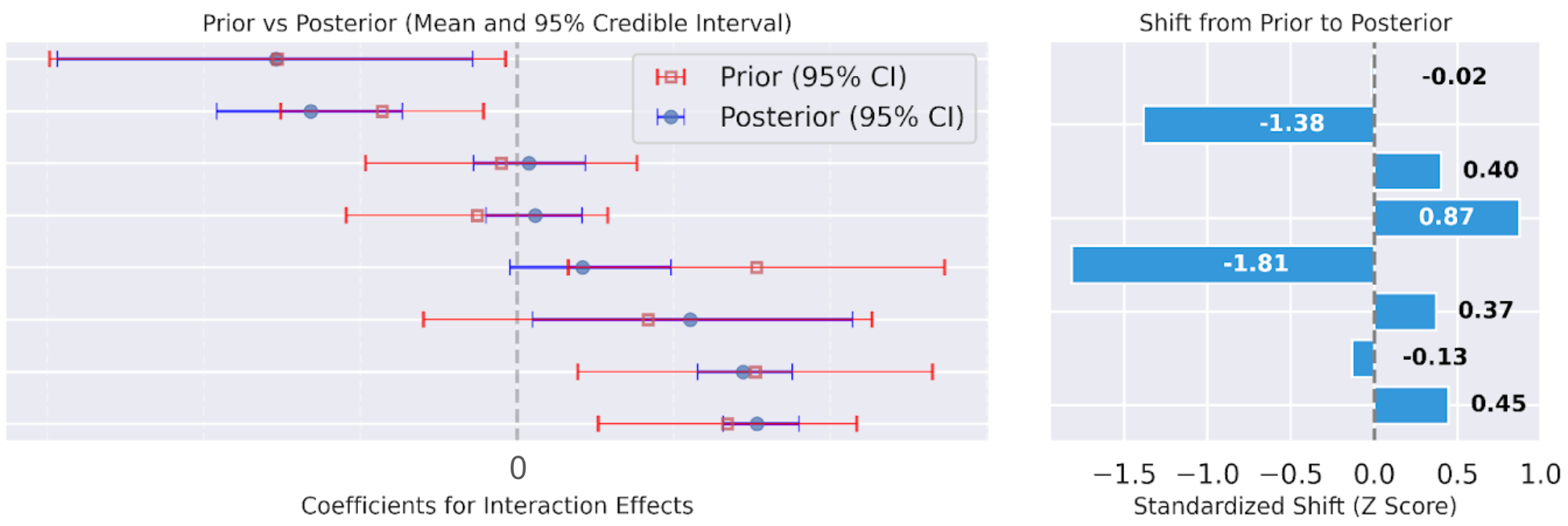}
  \label{fig:interaction}
\end{figure}

Figure~\ref{fig:interaction} illustrates the prior and posterior estimates for the most important interaction coefficients ($\lambda$) that determine heterogeneous incrementality estimates. We find that most of the posterior estimates are consistent with the prior, but more precise (a narrower 95\% credible interval compared to the 95\% confidence interval of the prior). Further, where the estimated posterior changes signs relative to the prior, the change makes sense.

\subsection{Comparing group-level supply incrementality: Prior vs Posterior}
\begin{figure}[h]
  \caption{Prior vs Posterior Estimates at Product Group Level}
  \centering
  \includegraphics[width=1\linewidth]{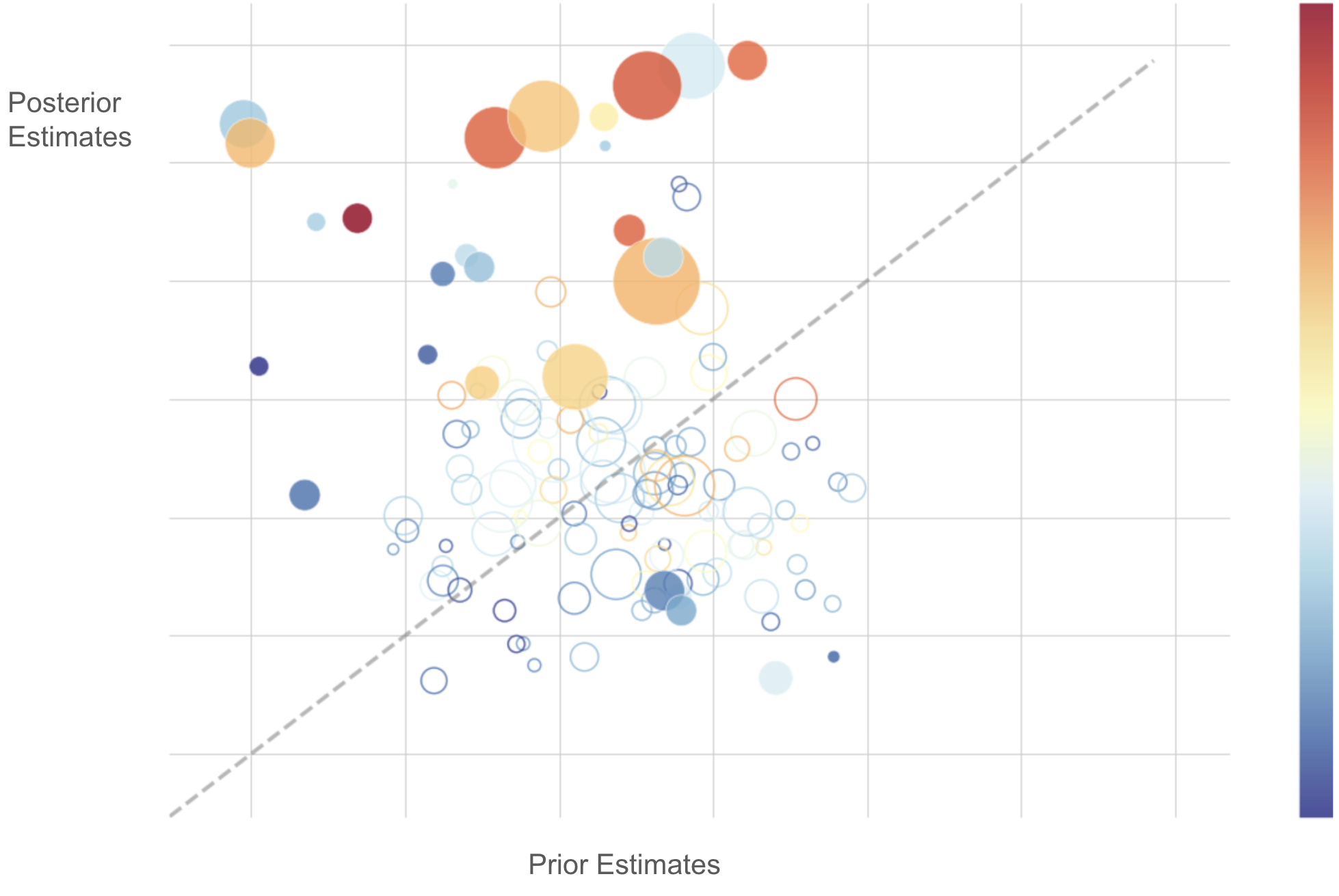}
  \label{fig:geoeffect}
\end{figure}

Figure~\ref{fig:geoeffect} compares the estimated supply incrementality (aggregated to the product group level, weighted by supply) vs. the prior estimate. In this figure, the size of the circle reflects the average supply level of the product group and the color of the circle reflects historical demand levels in the listing group.

This figure shows that in the largest groups where the posterior estimates differ significantly from the prior, the difference can often be explained by demand levels, suggesting that our model is capturing meaningful marketplace dynamics when substantial data is available to overcome the prior.

\subsection{Listing-segment level estimates}
Findings at the listing segment level are also corroborated with other research and domain knowledge. The pattern of listing segment-level incrementality aligns well with our understanding of marketplace dynamics, showing higher incrementality in areas with strong demand signals and unique listing attributes that are not easily substitutable. This validation through domain expertise gives us confidence in the model's ability to capture meaningful variations in the causal impact of supply across different market segments. The posterior estimates at the listing segment level provide scalable, consistent insights about supply incrementality that enable important business decisions to bring the most value to the community.    

\section{Model Evaluation and Future Work}
\subsection{Model Evaluation}
Model evaluation in our setting is particularly challenging due to the fundamental problem of causal inference - we never observe the counterfactual outcomes (\citet{sanchez2025airbnb};\citet{gelman2006data}). The ideal evaluation would compare our estimates to results from randomized experiments that randomize supply directly, but such experiments are typically infeasible in two-sided marketplaces. Instead, we focus on validating our model through consistency with theoretical expectations, alignment with domain expertise, and robustness to alternative specifications.

We validate the model output using the following approaches:
\begin{enumerate}
    \item \textbf{Listing group level validation}: We compare our listing group level estimates with the established prior work to ensure consistency at the group level. We leverage these existing estimates as priors and investigate cases where our posteriors deviate significantly from the priors to understand whether these differences are supported by additional data evidence.
    
    \item \textbf{Listing segment level validation}: We confirm that interaction coefficients are consistent with domain knowledge and marketplace intuition. For example, we validate that supply incrementality increases with or decreases in relevant features in a way that matches domain knowledge and marketplace intuition. We also conduct sanity checks on specific listing segments with extremely high or low estimated incrementality to ensure the results align with business understanding.

    \item \textbf{Natural experiments}: We aim to leverage potentially exogenous supply shocks, such as regulatory changes or natural events that affect supply in some areas but not others, as validation opportunities (\citet{baylis2019moral}).

\end{enumerate}

\section{Conclusion}
This paper introduces a novel approach to estimating supply incrementality in two-sided marketplaces with heterogeneous products by combining causal machine learning techniques with geospatial methods and Bayesian hierarchical modeling. Our methodology addresses several critical challenges in measuring the causal impact of additional supply on marketplace outcomes, particularly the endogeneity between supply and demand, substitution effects between similar products, and the need to account for pre-existing knowledge.

Beyond its application to Airbnb's marketplace, our methodology has broad relevance for other two-sided platforms facing similar challenges in understanding the incrementality of additional supply or products. For example, this approach could be adapted to ride-sharing platforms evaluating the impact of additional drivers, e-commerce marketplaces considering new sellers, or content platforms assessing additional creators.

Future research could extend this framework by incorporating temporal dynamics, exploring alternative methods for defining similarity between product segments, and developing experimental approaches to validate the causal estimates. As marketplace platforms continue to grow and diversify their offerings, methodologies that provide granular, causal insights into marketplace dynamics will become increasingly valuable for business decisions to bring more value to the community.

\section{Acknowledgement}
The authors would like to thank  Brian Weller (Airbnb, Inc.), Mitra Akhtari (Airbnb, Inc.), Totte Harinen (Airbnb, Inc.), Thomas Covert (Airbnb, Inc.), Carlos Sanchez-Martinez (Airbnb, Inc.), Yvonne Wang (Airbnb, Inc.), Allen Ross (Airbnb, Inc.), Shanni Weilert (Airbnb, Inc.),  Navin Sivanandam (Airbnb, Inc.) for their contributions and support for this project.

\nocite{berry1995auto}
 

\bibliography{sample-base}

\begin{thebibliography}{12}
\providecommand{\natexlab}[1]{#1}
\providecommand{\url}[1]{\texttt{#1}}
\expandafter\ifx\csname urlstyle\endcsname\relax
  \providecommand{\doi}[1]{doi: #1}\else
  \providecommand{\doi}{doi: \begingroup \urlstyle{rm}\Url}\fi

\bibitem[Ye et~al.(2018)Ye, Qian, Chen, Wu, Zhou, De~Mars, Yang, and Zhang]{ye2018customized}
Peng Ye, Julian Qian, Jieying Chen, Chen-hung Wu, Yitong Zhou, Spencer De~Mars, Frank Yang, and Li~Zhang.
\newblock Customized regression model for {Airbnb} dynamic pricing.
\newblock In \emph{Proceedings of the 24th ACM SIGKDD International Conference on Knowledge Discovery \& Data Mining}, pages 932--940, New York, NY, USA, 2018. Association for Computing Machinery.

\bibitem[Sanchez~Martinez et~al.(2025)Sanchez~Martinez, O'Donnell, Yuan, and Zhu]{sanchez2025airbnb}
C.~Sanchez~Martinez, S.~O'Donnell, L.~Yuan, and Y.~Zhu.
\newblock How {Airbnb} measures listing lifetime value.
\newblock \emph{the Airbnb Tech Blog}, 2025.

\bibitem[Douglas(1976)]{douglas1976cobb}
Paul~H. Douglas.
\newblock The {Cobb-Douglas} production function once again: its history, its testing, and some new empirical values.
\newblock \emph{Journal of Political Economy}, 84\penalty0 (5):\penalty0 903--915, 1976.

\bibitem[Petrongolo and Pissarides(2001)]{petrongolo2001looking}
Barbara Petrongolo and Christopher~A. Pissarides.
\newblock Looking into the black box: A survey of the matching function.
\newblock \emph{Journal of Economic Literature}, 39\penalty0 (2):\penalty0 390--431, 2001.

\bibitem[Gandhi and Nevo(2021)]{gandhi2021empirical}
Amit Gandhi and Aviv Nevo.
\newblock Empirical models of demand and supply in differentiated products industries.
\newblock Nber working paper, National Bureau of Economic Research, 2021.

\bibitem[Berry et~al.(1995)Berry, Levinsohn, and Pakes]{berry1995automobile}
Stephen Berry, James Levinsohn, and Ariel Pakes.
\newblock Automobile prices in market equilibrium.
\newblock \emph{Econometrica}, 63\penalty0 (4):\penalty0 841--890, 1995.

\bibitem[Chernozhukov et~al.(2018)Chernozhukov, Chetverikov, Demirer, Duflo, Hansen, Newey, and Robins]{chernozhukov2018double}
Victor Chernozhukov, Denis Chetverikov, Mert Demirer, Esther Duflo, Christian Hansen, Whitney Newey, and James Robins.
\newblock Double/debiased machine learning for treatment and structural parameters.
\newblock \emph{The Econometrics Journal}, 21\penalty0 (1):\penalty0 C1--C68, 2018.

\bibitem[Kennedy(2022)]{kennedy2022towards}
Edward~H. Kennedy.
\newblock Towards optimal doubly robust estimation of heterogeneous causal effects, 2022.

\bibitem[Bergmeir et~al.(2018)Bergmeir, Hyndman, and Koo]{bergmeir2018note}
Christoph Bergmeir, Rob~J. Hyndman, and Bonsoo Koo.
\newblock A note on the validity of cross-validation for evaluating autoregressive time series prediction.
\newblock \emph{Computational Statistics \& Data Analysis}, 120:\penalty0 70--83, 2018.

\bibitem[Credit and Lehnert(2024)]{credit2024structured}
Kevin Credit and Matthias Lehnert.
\newblock A structured comparison of causal machine learning methods to assess heterogeneous treatment effects in spatial data.
\newblock \emph{Journal of Geographical Systems}, 26:\penalty0 483--510, 2024.

\bibitem[Gelman and Hill(2006)]{gelman2006data}
Andrew Gelman and Jennifer Hill.
\newblock \emph{Data Analysis Using Regression and Multilevel/Hierarchical Models}.
\newblock Cambridge University Press, 2006.

\bibitem[Baylis and Boomhower(2019)]{baylis2019moral}
Patrick Baylis and Judson Boomhower.
\newblock Moral hazard, wildfires, and the economic incidence of natural disasters.
\newblock NBER Working Paper 26550, National Bureau of Economic Research, 2019.

\end{thebibliography}


\end{document}